%% file: main.tex
\documentclass[conference]{IEEEtran}

\IEEEoverridecommandlockouts
\usepackage{cite}
\usepackage{amsmath,amssymb,amsfonts}
\usepackage{algorithmic}
\usepackage{float}
\usepackage{graphicx}
\usepackage{enumitem}
\usepackage{caption}
\usepackage{subfloat}
\usepackage{subcaption}
\usepackage{textcomp}
\usepackage{xcolor}
\usepackage{booktabs,siunitx,tabularx, stfloats}
\usepackage{threeparttable}
\usepackage{svg}
\usepackage{diagbox}
\usepackage{fancyhdr}
\usepackage{bbding}
\renewcommand{\baselinestretch}{0.9074}

\def\BibTeX{{\rm B\kern-.05em{\sc i\kern-.025em b}\kern-.08em
    T\kern-.1667em\lower.7ex\hbox{E}\kern-.125emX}}
\newcommand{\printfnsymbol}[1]{%
  \textsuperscript{\@fnsymbol{#1}}%
}

\newcommand{\I}{\mathbf{I}}
\newcommand{\Patch}{\mathbf{P}}

\usepackage{glossaries}
\newacronym{gan}{GAN}{Generative Adversarial Network}
\newacronym{da}{DA}{Data Augmentation}
\newacronym{nn}{NN}{Neural Network}
\newacronym{cnn}{CNN}{Convolutional Neural Network}
\newacronym{cv}{CV}{Computer Vision}
\newacronym{sd}{SD}{Stable Diffusion}
\newacronym{sde}{SDE}{Stochastic Differential Equation}
\newacronym{ldm}{LDM}{Latent Diffusion Model}
\newacronym{dm}{DM}{Diffusion Model}
\newacronym{csam}{CSAM}{Child Sexual Abuse Material}
\newacronym{auc}{AUC}{Area Under the Curve}
\newacronym{ba}{B-ACC}{Balanced Accuracy}
\newacronym{ba0}{$\textrm{B-ACC}_{@ \textrm{thr}=0}$}{Balanced Accuracy at threshold 0}
\newacronym{tpr0}{$\textrm{TPR}_{@ \textrm{thr}=0}$}{True Positive Rate at threshold 0}
\newacronym{fpr0}{$\textrm{FPR}_{@ \textrm{thr}=0}$}{False Positive Rate at threshold 0}

\begin{document}

\sloppy

\title{When Synthetic Traces Hide Real Content: \\ Analysis of Stable Diffusion Image Laundering}

\author{Sara Mandelli,  Paolo Bestagini, Stefano Tubaro\\ 
\small{Dipartimento di Elettronica, Informazione e Bioingegneria, Politecnico di Milano, 20133 Milan, Italy.} \\
\small{Emails: name.surname@polimi.it}\\
\thanks{This research is sponsored by the Defense Advanced
Research Projects Agency (DARPA) and the Air Force Research Laboratory
(AFRL) under agreement number FA8750-20-2-1004. 
This work
was partially supported by the European Union under the Italian National
Recovery and Resilience Plan (NRRP) of NextGenerationEU
(PE00000001 - program ``RESTART'' and PE00000014 - program ``SERICS'') and 
by the ``FOSTERER'' project, funded by the Italian Ministry of University and Research within PRIN 2022 program.}
\vspace{-5pt}
}
\maketitle

\begin{abstract}
In recent years, methods for producing highly realistic synthetic images have significantly advanced, allowing the creation of high-quality images from text prompts that describe the desired content. Even more impressively, Stable Diffusion (SD) models now provide users with the option of creating synthetic images in an image-to-image translation fashion, modifying images in the latent space of advanced autoencoders. This striking evolution, however, brings an alarming consequence: it is possible to pass an image through SD autoencoders to reproduce a synthetic copy of the image with high realism and almost no visual artifacts. This process, known as SD image \textit{laundering}, can transform real images into lookalike synthetic ones and risks complicating forensic analysis for content authenticity verification. Our paper investigates the forensic implications of image laundering, revealing a serious potential to obscure traces of real content, including sensitive and harmful materials that could be mistakenly classified as synthetic, thereby undermining the protection of individuals depicted. To address this issue, we propose a two-stage detection pipeline that effectively differentiates between pristine, laundered, and fully synthetic images (those generated from text prompts), showing robustness across various conditions. Finally, we highlight another alarming property of image laundering, which appears to mask the unique artifacts exploited by forensic detectors to solve the camera model identification task, strongly undermining their performance. Our experimental code is available at
\\ https://github.com/polimi-ispl/synthetic-image-detection.
\end{abstract}


\vspace{-10pt}
\input{intro}

\input{problem}

\input{method}

\input{experiments}
\input{camera_model_problem}

\input{conclusions}

\bibliographystyle{IEEEtran}
\bibliography{references}

\end{document}

%% file: intro.tex
\section{Introduction}
\label{sec:intro}

Over the last few years, we have witnessed an escalation in methods for producing increasingly realistic synthetically generated images, which exhibit high quality and realism, easily fooling the human eye~\cite{Karras2021, Rombach2022ldm, dalle3, firefly, imagine}. In particular, \glspl{dm}~\cite{Dickstein2015dm_init} have been dominating the scene due to continuous improvements that have brought astonishing generation results~\cite{ho2020ddpm, scorebased, dhariwal2021diffusionbeatgans}. 
\glspl{ldm}~\cite{Rombach2022ldm} have been lately introduced to enhance the visual fidelity of generated images and reduce the training complexity compared to standard \glspl{dm}. These models allow to generate synthetic images starting from a noisy signal and a text prompt that describes the desired characteristics for the final generated image. Both input signal and text are encoded into latent representations using powerful pretrained autoencoders. Subsequent denoising steps in the latent space, followed by a decoding stage, produce the final synthetic image.

More recently, \glspl{ldm} have evolved into \gls{sd} models, thanks to even more advanced autoencoder structures trained on vast amounts of data~\cite{sd1_github, sd2_github, podell2023sdxl, sauer2023adversarial-sdxl-turbo}. \gls{sd} has expanded the options available to users, allowing them to build creations starting from actual input images. Indeed, \gls{sd} enables users to provide an image as input to the generation pipeline. This image is superimposed with noise, encoded into the latent space and modified through multiple denoising steps influenced by the user's text prompt instructions. The similarity of the final decoded image to the input one can be adjusted by the user using a specific strength parameter.

\begin{figure}[t]
        \centering
    \includegraphics[width=\columnwidth]{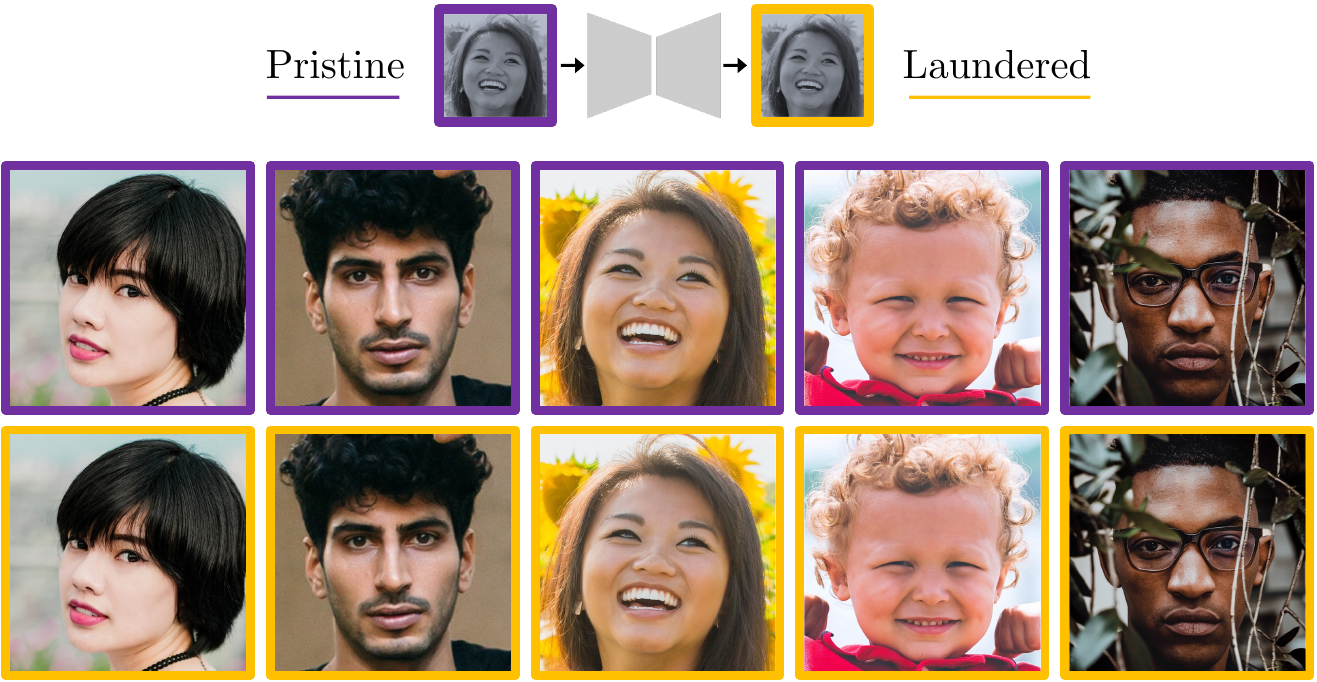}
        \caption{Pristine images and their laundered copies obtained by passing pristine samples through SD autoencoder. Laundered samples look extremely realistic, with almost a total absence of notable generation artifacts, even in the case of uncommon patterns that could be harder to reproduce. }
        \label{fig:teaser}
\vspace{-20pt}
\end{figure}

At one extreme, the user can adjust this strength parameter to select the maximum similarity between the input and output images. As a result, the input image will be encoded in the latent space without any noise injections; the latent representation will not undergo any denoising steps, and it will be directly decoded to produce the final image. This ``generated'' image will almost perfectly reproduce the semantic content of the input one, with very few details, barely visible to the human eye, that can change according to the specific autoencoder architecture.

This process can be applied to any input image, including real content like photographs or video frames. In the forensic community, this technique has come to be known as image \textit{laundering}, meaning that real content can be transformed through a chain of \gls{sd} encoding and decoding to wash out its original traces and simulate synthetic generation~\cite{cozzolino2024raising}. Some examples of real photographs and their laundered versions are shown in Fig.~\ref{fig:teaser}. Notice the extremely high realism of the laundered copies and the absence of noticeable artifacts upon visual inspection. However, in principle, laundered images are synthetic and not real photographs.

This paper proposes a forensic analysis of the image laundering process, demonstrating that it represents a serious issue for assessing content authenticity on the web. In fact, the laundered copies of real images carry common artifacts of synthetically generated content, since they are passed through the autoencoding process typical of \gls{sd} generation. Even the most advanced forensic detectors developed to separate real and synthetic images are at high risk of being fooled~\cite{cozzolino2024raising}, as they will likely categorize the laundered images as synthetic.

Suppose a malicious user captures photographs depicting sensitive content, including extreme cases such as pornographic images or \gls{csam}. In our experiments, we demonstrate that image laundering provides a concrete method to conceal the user traces, masking the images as if they were generated by \gls{sd}. Investigators or content moderators relying on synthetic image detectors might mistakenly believe the images to be synthetic and thus lower their guard when assessing content spread across the web or social media. This scenario carries severe consequences, including the potential dissemination of original sensitive content, compromising individuals protection. Sensitive or harmful materials risk being overlooked or not given appropriate attention, increasing the chance of perpetuating online.


In our paper, we analyze in depth the image laundering operation and highlight the issues it presents for forensic investigators. Specifically, we demonstrate that forensic detectors developed to distinguish real from synthetic content are inadequate for classifying laundered images. We then analyze the frequency spectrum of laundered images in comparison to their real counterparts and to fully synthetic content (images generated from input noise and text prompts).

Furthermore, we propose a straightforward solution to address this issue: a two-stage detection pipeline that effectively differentiates between real, laundered, and fully synthetic images. Our solution generalizes well across different semantic contents and unseen data during the training stage, and proves robust to common post-processing operations such as compression and resizing.

In a final experiment, we reveal another alarming consequence of image laundering: our results show that laundering significantly undermines state-of-the-art detectors developed for the camera model identification task. This operation appears to wash out the footprints of the specific camera model used to capture a photograph, effectively concealing the subtle artifacts that allow investigators to trace back its origin.

%% file: problem.tex
\section{Stable Diffusion and Image Laundering Issue}
\label{sec:problem}

\subsection{Image-to-image synthesis through Stable Diffusion}

In a ``standard'' text-to-image generation through \glspl{ldm}, the model is fed with random noise~\cite{Rombach2022ldm}. This input signal can be modeled as a noise-free image super-imposed to a strong noise which hides the scene content depicted on it. 
The provided text prompt acts as a hint for what the original image (before noise corruption) should look like. 
To produce the noise-free image, the \gls{ldm} converts the noise into a latent representation through an encoder and performs several denoising steps in this latent space.
This process is completed with cross-attention mechanisms that allow a latent representation of the text prompt to influence the denoising stage~\cite{Rombach2022ldm}. The process runs iteratively for a number of steps, such to gradually remove a portion of the noise at a time. The final image is estimated by decoding the output of multiple denoising steps.    

In the last period, \glspl{ldm} have started being known as \gls{sd} models. This happened after considering advanced image autoencoders and text prompt encoders trained on a huge amount of data~\cite{sd1_github}. 
\gls{sd} allows to generate images from a text prompt, but it allows as well to synthesize images in an image-to-image translation fashion. 
In detail, image-to-image emulates the standard image synthetis process, apart from the fact that the input image is no more a random noise but it is an actual image, like a real photograph or a cartoon illustration~\cite{sd1_github}. 
To generate a synthetic image, the input original image is superimposed to an additive noise term to generate a noisy version of it. This noisy sample is passed through \gls{sd} generation and the output is another image that resembles more closely the original supplied one with respect to the random noise input case. 

To control how much the output is affected by the input, \gls{sd} is provided with a ``strength'' parameter $s \in [0, 1]$. 
This parameter is related to the amount of noise added at the beginning, and to the number of denoising steps that \gls{sd} will run. Higher values of $s$ will deviate more the output image from the input original one. 

\subsection{Image laundering through Stable Diffusion}
\label{subsec:problem_laundering}

In the image-to-image synthesis process, the strength parameter $s$ has a considerable impact on the semantic consistency between the input and the output images.  
In this context, $s = 1$ and $s = 0$ describe two extreme scenarios of the possible generation outcomes. 
In case $s$ is $1$, the noise addition step does not completely destroy the input image,
even if this allows for lots of variations and semantics different from the input one. 
In case of $s = 0$, no noise is added to the input image and no denoising steps are run. In this setup, the text prompt is completely irrelevant for the final generation. The image is passed only through \gls{sd} encoder and decoder, and the output image is highly connected to the original one. The semantic content is completely replicated, and very few details (barely visible at the human eye) can be changed, according to the specific autoencoder used (see some examples in Fig.~\ref{fig:teaser}). 

This last procedure can be applied to any input image, including real content like photographs or video frames. When applied to real images, this process has started been known as image \textit{laundering}, meaning that real content can be transformed through a chain of \gls{sd} encoding and decoding to wash out its original traces and simulate a synthetic generation~\cite{cozzolino2024raising}. 
The available synthetic image detectors in the state-of-the-art risks to be fooled by this process, detecting the laundered images as being synthetic~\cite{cozzolino2024raising}. 
As a result, original images showing sensitive or harmless content could proliferate online without being identified as real, risking oversight and underming individuals protection.
In the next lines, we detail the proposed solution to deal with such alarming problem.

%% file: method.tex
\section{Image Laundering Detection}
\label{sec:method}

\subsection{Problem formulation}
\label{subsec:method_problem}


Given an image $\I$ under analysis, we aim at investigating whether $\I$ can be correctly classified as being real, fully synthetic or the result of a laundering operation applied to a real image. 
For clarity's sake, we define an image $\I$ to be ``real'' if its pixel content generally comes from a photograph; it can have undergone post-processing operations like compression, cropping or resizing, but its original content has been acquired by a digital camera sensor.
On the contrary, we say that $\I$ is ``synthetic'' if it is the result of a synthetic generation model applied to an input signal. 

If the input signal is a random noise,
we define the generated image to be ``fully synthetic''; for instance, \gls{gan}-generated images and \gls{sd} text-to-image
belong to the ``fully synthetic'' category~\cite{Karras2021, Rombach2022ldm}. 
If the input signal is a real image passed through \gls{sd} encoding and decoding with strength $s=0$ described in Section~\ref{subsec:problem_laundering}, we define the image to be ``laundered''. 
For simplicity, we do not explore the intermediate scenario of image-to-image translation with strength $s > 0$. To our knowledge, this is the first study exploring the issue of laundered image detection, thus we do not want to complicate our analysis, leaving the rest for future investigations. 

Our goal is two-fold: first, we investigate whether laundered images can be differentiated from fully synthetic ones by existing forensic detectors developed to deal with the real versus synthetic detection task. If this is not the case, we investigate whether there are any traces that can be exploited to tell fully-synthetic images and laundered images apart.
In doing so, we are interested in building a detection pipeline that can work with the three categories of images (real, synthetic and laundered) at the same time.

\subsection{Forensics analysis of laundered images}
\label{subsec:method_method}

\textbf{Backbone detector. }
To perform our investigations, we exploit a detector built upon the one proposed in~\cite{mandelli2022detecting}, which has shown excellent performances for the real versus synthetic image detection task. This detector is based on the extraction of small squared patches from the query image and their subsequent aggregation to assign a single score per image. It proved very robust to compression and resizing operations, thanks to a long list of augmentations included in the training process. Furthermore, the possibility of extracting small patches from the query image enables to be less dependent on the semantic content depicted in the image and to focus on the actual synthetic generation artifacts.

\begin{figure}[t]
        \centering
        \includegraphics[width=\columnwidth]{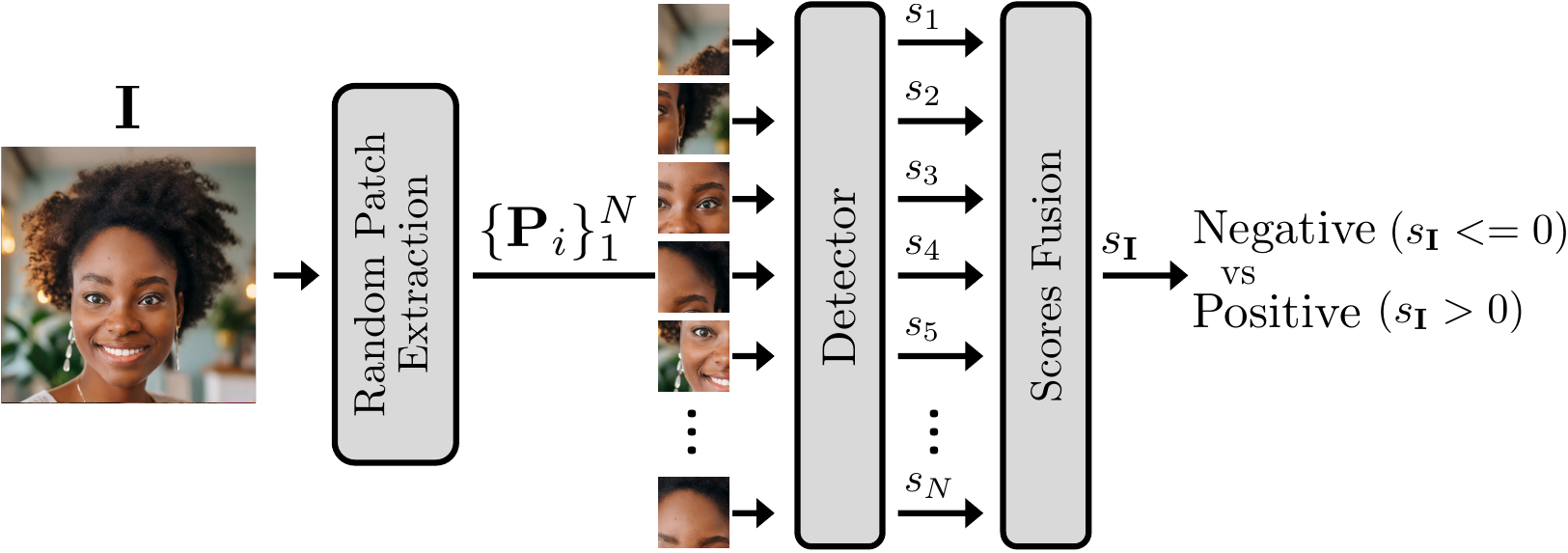}
        \caption{Scheme of the proposed backbone detector based on the random extraction of $N$ squared patches from the input image. }
        \label{fig:generic_detector_patches}
\vspace{-15pt}
\end{figure}

A sketch of our proposed detector is shown in Fig.~\ref{fig:generic_detector_patches}.
As done in~\cite{mandelli2022detecting}, we consider the EfficientNet-B4 \gls{cnn} architecture~\cite{tan2019efficientnet}, but we further simplify the patch extraction and aggregation strategy. Differently from~\cite{mandelli2022detecting}, we always extract in random locations $N = 800$ color patches $\{\Patch_i\}_1^N$ with size $96 \times 96$ pixels, independently on the input image size. These numbers have been selected to enable good robustness and generalization properties even on large input images ($\sim 3000 \times 3000$ pixels). Every patch $\Patch_i$ is associated with a detection score $s_i$, which is greater than $0$ if the patch is detected as positive (i.e., ``synthetic''), and negative (i.e., ``real'') otherwise.  
After analyzing each patch through our detector, we aggregate the patches' scores $\{s_i\}_i^N$ by simply selecting the $M$ highest scores corresponding to the uppermost $75\%$ (in this case, $M = 600$) and computing their arithmetic mean. The final image score is defined as $s_{\mathbf{I}} = \sum_{i}^M s_i /M $.

\textbf{Preliminary results on laundered images. }
To motivate the tackled task, we perform some preliminary experiments by testing the above described detector over an unseen dataset of 1K real human faces~\cite{kaggle-real} and their related laundered versions generated by us through different \gls{sd} releases, namely \gls{sd}-1.5~\cite{sd1_github}, \gls{sd}-2.1~\cite{sd2_github}, \gls{sd}-XL~\cite{podell2023sdxl} and \gls{sd}-XL-turbo~\cite{sauer2023adversarial-sdxl-turbo}. 
In specific, we test a detector that was not trained over any laundered images, but was trained to discriminate between real and fully synthetic human faces generated through \glspl{gan} and \glspl{dm} (more details on the training setup in Section~\ref{sec:experiments}). 

\begin{figure}[t]
\centering
        \includegraphics[width=.9\columnwidth]{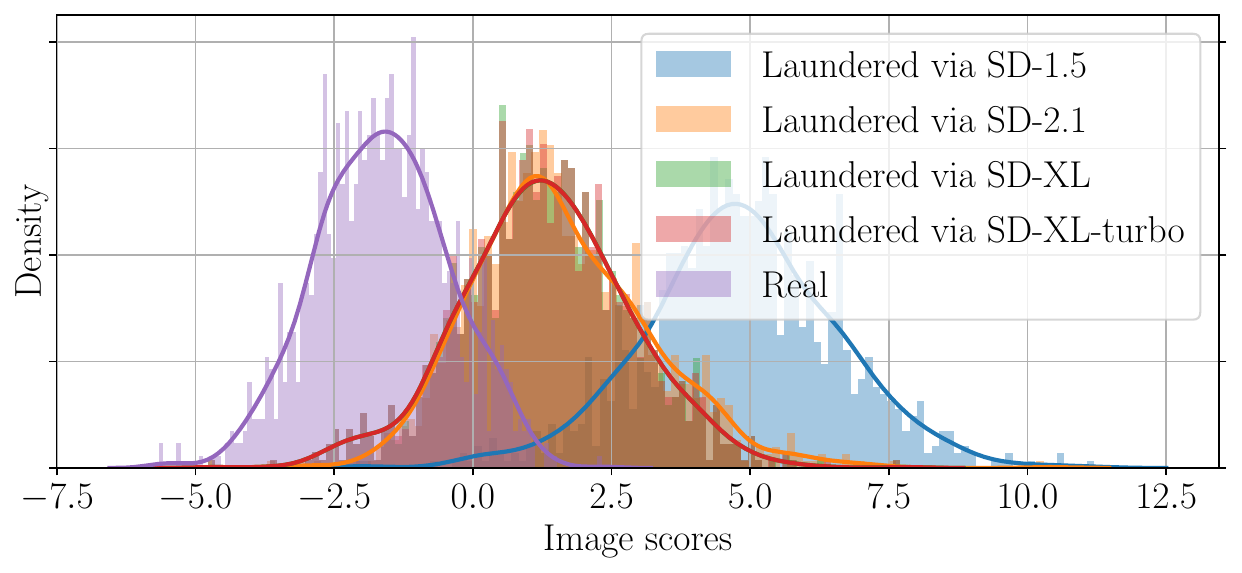}
        \caption{Distributions of the detection scores of real and laundered images; the detector is a synthetic versus real detector not trained over laundered images. }
\vspace{-10pt}
\label{fig:preliminary_results}
\end{figure}

Images' detection scores computed with the previously reported pipeline are shown in Fig.~\ref{fig:preliminary_results}. 
Though the laundering operation was unknown at training stage, the strongest majority of laundered images is detected as being synthetic (i.e., scores are greater than $0$). 
These results are quite alarming: in a realistic scenario, laundered versions of pristine images carrying sensitive content would likely be labeled as fake, with the risk of content moderators lowering down their guard and the potential spreading of real harmful material online. 


\begin{figure}[t]
\centering
    \begin{subfigure}[b]{\columnwidth}
        \centering
        \includegraphics[width=\columnwidth]{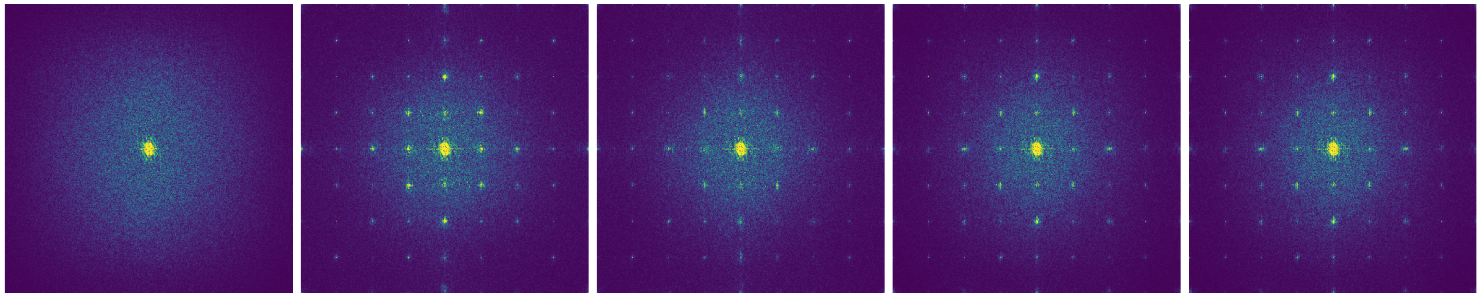}
        \caption{\scriptsize{Average Fourier spectra (magnitude) of real images and their laundered versions through SD-1.5, SD-2.1, SD-XL and SD-XL-turbo, respectively.}}
        \label{fig:fft_analysis_real_i2i}
    \end{subfigure}
\hfill
    \begin{subfigure}[b]{\columnwidth}
        \centering
        \includegraphics[width=.8\columnwidth]{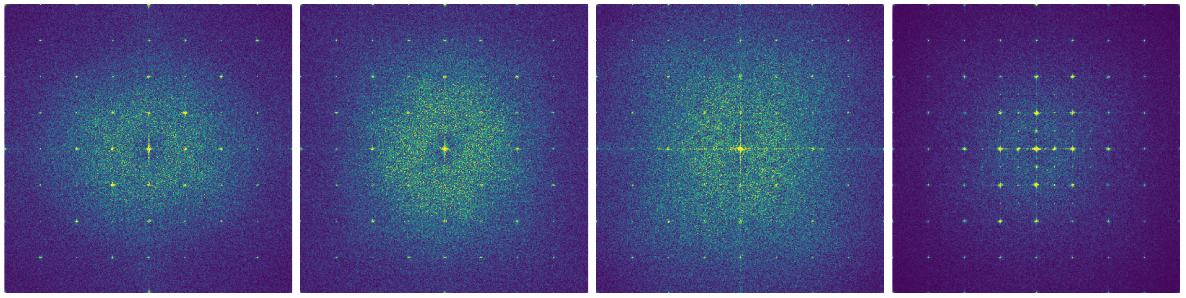}
        \caption{\scriptsize{Average Fourier spectra (magnitude) of fully synthetic images generated from text prompts through SD-1.5, SD-2.1, SD-XL and SD-XL-turbo, respectively.}}
        \label{fig:fft_analysis_t2i}
    \end{subfigure}
\caption{Fourier transform analysis of real, laundered and fully synthetic images. All spectra are centered in the spatial frequencies $(0, 0)$.
}   
\label{fig:fft_analysis}
\vspace{-15pt}
\end{figure}

\textbf{Frequency analysis. }
We deepen our investigations and we explore if there are any traces that can help differentiating laundered images from fully synthetic ones. To this purpose, we follow a well known strategy in the forensic community, i.e., we compute the average Fourier spectrum from noise residuals extracted from the images~\cite{cozzolino2024raising, vahdati2024beyond}. 
For this experiment, we consider the same set of real and laundered images used above, enlarged to include fully synthetic human faces generated via text prompts through the same generators. We extract noise residuals with the help of a standard denoiser~\cite{zhang2017beyond} which has been employed several times for the same task~\cite{corvi2023detection}. 

Fig.~\ref{fig:fft_analysis} depicts the average Fourier spectra of noise residuals. Real images are the only ones without any peaks in their Fourier transform, and this was expected~\cite{cozzolino2024raising}. More interestingly, laundered images show a different pattern with respect to fully synthetic images generated through the same generators. While fully synthetic samples present in general a relatively spread spectrum (except for \gls{sd}-XL-turbo), laundered samples show an extremely similar spectrum to the real images, which is more focused around lower frequencies,  except for the typical peaks caused by resampling operations in \gls{sd} autoencoders. 

These spectral differences suggest that there is a concrete possibility to distinguish laundered from fully synthetic samples. In the next lines, we propose a simple yet effective solution to discriminate between them.


\begin{figure*}[t]
        \centering
        \includegraphics[width=.7\textwidth]{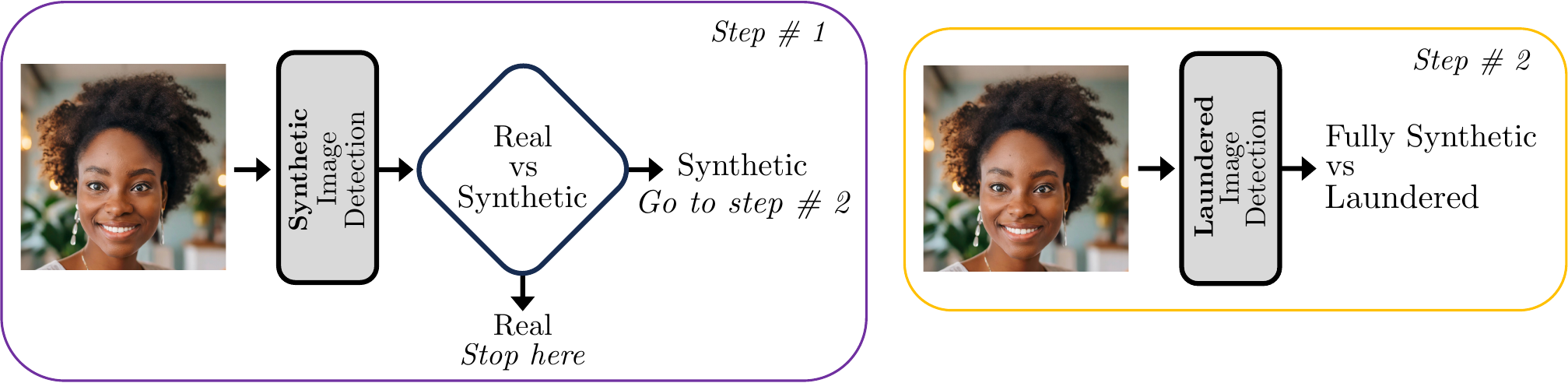}
        \caption{Sketch of our proposed two-stage methodology to classify an analyzed image as real, fully synthetic or laundered. At step 1, we tell real images apart from synthetic ones; step 2 further discriminates between fully synthetic and laundered samples.}
        \label{fig:method}
\vspace{-15pt}
\end{figure*}

\textbf{Proposed two-stage detection. }
To deal with the image laundering issue, we propose a detection pipeline consisting of two subsequent steps (see Fig.~\ref{fig:method}):
\begin{itemize}
    \item \textit{Synthetic image detection}, where we distinguish between pristine and synthetic samples; synthetic samples involve both fully synthetic and laundered images.
    \item \textit{Laundered image detection}, where we differentiate between fully synthetic images and laundered ones.
\end{itemize} 

We employ two detectors with the same backbone structure described before, operating on a patch-wise basis, which makes our methodology less dependent on the image content. The first detector is trained over real and synthetic samples; the second includes only synthetic samples in its training. In the last case, the positive class describes the laundered samples, while the negative one refers to fully synthetic samples. 

At deployment stage, we pass the query image through the first detector. If the image score is less then $0$, the image is detected as being real and we stop our analysis. Otherwise, the image is passed through the second detector, which classifies fully synthetic and laundered images apart. 
In our experimental campaign, we prove that this straightforward methodology is effective to keep at bay the laundering problem, being robust to post-processing operations like compression and resampling.

%% file: experiments.tex
\section{Experimental Analysis}
\label{sec:experiments}


\subsection{Datasets}
\label{subsec:datasets}

\textbf{Training datasets. } 
The first detector (real versus synthetic) is trained over real and synthetic human faces collected from several state-of-the-art datasets. We train over a huge amount of data to ensure good robustness and generalization against different data and synthetic generators. Overall, the training set includes more than $200$K images (equally balanced among real and fake samples) with minimum size $256 \times 256$ pixels. 

Real faces have been selected from FFHQ~\cite{ffhq}, CelebAHQ~\cite{celebahq} and from the pristine dataset released in~\cite{wang2020cnn}. 

Synthetic faces have been generated through state-of-the-art models for synthetic image generation. These include \gls{gan}-based generators like StyleGAN2~\cite{sg2}, StyleGAN3~\cite{Karras2021}, StarGAN-v2 \cite{afhq}, FaceVid2Vid \cite{facev2v} and Taming Transformers \cite{taming_tx}, as well as \gls{dm}-based generators like Score-based models~\cite{scorebased}, \glspl{ldm} and \gls{sd} models~\cite{Rombach2022ldm}, DALL$\cdot$E\,3-based Image Creator~\cite{dalle3}, Adobe Firefly~\cite{firefly} and Imagine from Meta AI~\cite{imagine}.

We use images from \gls{sd}-1.5, \gls{sd}-2.1, \gls{sd}-XL and \gls{sd}-XL-turbo in both laundered and text-prompt modes; the laundered images are synthetic copies of the considered real data, while text-prompt mode includes only images generated from noise (what we defined to be ``fully synthetic''). Notice that, in the preliminary experiments shown in Section~\ref{sec:method}, laundered images were not included during training.


The second detector (fully synthetic versus laundered) is trained over synthetic images only, including all the synthetic data used at the first training stage. 

\textbf{Test dataset. } 
We evaluate results on real and synthetic images selected from different datasets than training ones. 

Pristine images belong to the dataset presented in~\cite{kaggle-real}, consisting of $1081$ real human faces with size $600 \times 600$ pixels.

Synthetic images are of three different categories: (i) laundered versions of pristine data computed through \gls{sd}-1.5, \gls{sd}-2.1, \gls{sd}-XL and \gls{sd}-XL-turbo ($1081$ images each); (ii) fully synthetic faces generated from text prompts through the same generators ($\sim 300$ images per generator); (iii) $3000$ uncontrolled synthetic data selected from the DiffusionDB dataset~\cite{diffusion_db}, which was built by scraping user-generated images on the official \gls{sd} Discord server and depicts various semantic contents. This last dataset is uncontrolled, since we do not have precise information on the image generation process, i.e., images could have been laundered or generated from images or noise via text prompts. To be sure of testing realistic samples resembling actual photographs, DiffusionDB data have been filtered out by removing cartoon-like samples.

\subsection{Training details}
\label{subsec:experiments_training}

Following the approach in~\cite{mandelli2022detecting}, we initialize the network weights with those trained on the ImageNet database. We employ cross-entropy loss and the Adam optimizer with default parameters, training for up to 500 epochs. The learning rate starts at 0.001 and is reduced by a factor of 10 if the loss does not decrease for 10 epochs. Training is halted if the validation loss fails to improve for over 20 epochs, and the model with the best validation loss is selected.
To enhance robustness and generalization, we include a consistent amount of training data augmentations, including random resizing, compression and color corrections as suggested in~\cite{mandelli2022detecting}.

\subsection{Experimental results}
\label{sec:results}

\begin{figure}[t]
\centering
        \includegraphics[width=.9\columnwidth]{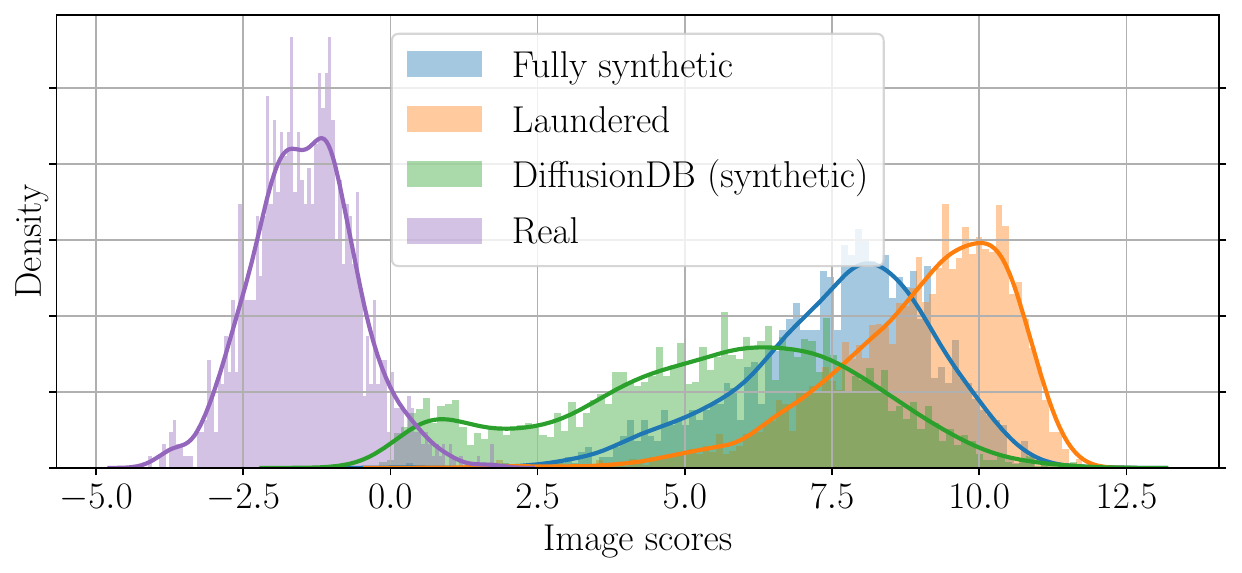}
        \caption{Distributions of the detection scores of the first detection stage. }
\label{fig:detector1_results}
\vspace{-5pt}
\end{figure}

\textbf{Synthetic image detection. }
In this phase, we pass the entire test set through our real vs synthetic detector. The achieved image scores distributions are depicted in Fig.~\ref{fig:detector1_results}. 

It is worth noticing that the \gls{tpr0} is extremely high for both laundered and fully synthetic samples, being different from $100\%$ only for DiffusionDB data (in this case, we achieve $99.67\%$). 
As previously conjectured, the detector proves robust to semantic content different from the training one.
Moreover, we report a reduced \gls{fpr0} $= 5.74\%$, meaning for good generalization 
on unseen pristine data at training stage. 


\textbf{Laundered image detection. }
We now pass only the synthetic test set throughout our fully synthetic vs laundered detector. 
We do not include anymore the pristine samples, since we want to evaluate the detection performances at the best conditions ever, i.e., supposing all pristine samples have been already filtered out by the first detection step. 
Moreover, we do not consider images of DiffusionDB dataset, since we do not have complete control on their generation process (i.e., they could have been laundered or generated from images with strength $s > 0$ or generated from text prompts).  

The first column of Table~\ref{tab:second detection stage} reports the evaluation metrics achieved. 
We obtain excellent performances: the \gls{auc} is above $0.99$ and the maximum \gls{ba} exceeds $96\%$. 
Moreover, the maximum \gls{ba} and that evaluated at threshold $0$ (\gls{ba0}) are very close one with the other, meaning for almost no need of calibrating our detector on unseen data. 

We deepen investigations by comparing laundered and fully synthetic samples generated through one different \gls{sd} version at a time. The achieved detection results are shown in Table~\ref{tab:second detection stage}, starting from second column. Similarly to the full dataset (first column), we achieve optimal performances for all \gls{sd} releases. Insterestingly, the model carrying the strongest laundering artifacts is \gls{sd}-XL-turbo, which is the most up-to-date \gls{sd} version out of the investigated ones, thus it might represent the chosen \gls{sd} option by users in the near future.

\begin{table}[t]
\centering
\caption{Achieved results at the second detection stage of fully synthetic vs laundered. In bold, the best results per metrics. }
\label{tab:second detection stage}
\resizebox{\columnwidth}{!}{
\begin{tabular}{@{}lccccc@{}}
\toprule
 & All data & SD-1.5 & SD-2.1 & SD-XL & SD-XL-turbo \\
\midrule
AUC &$0.994$ & $0.993$ & $0.994$ & $0.993$ & $\mathbf{0.999}$ \\
$\textrm{B-ACC}_{\max}$ & $96.1\%$  & $95.8\%$ & $96.6\%$ & $96.9\%$ & $\mathbf{99.3\%}$ \\
$\textrm{B-ACC}_{@ \textrm{thr}=0}$ & $95.9\%$ & $93.9\%$ & $96.2\%$ & $94.8\%$ & $\mathbf{98.6\%}$ \\
$\textrm{TPR}_{@ \textrm{thr}=0}$ & $94.8\%$ & $89.8\%$ & $94.9\%$ & $97.1\%$ & $\mathbf{97.1\%}$ \\
$\textrm{FPR}_{@ \textrm{thr}=0}$ & $2.9\%$ & $2.0\%$ & $2.5\%$ & $7.4\%$ & $\mathbf{0.0\%}$ \\
\bottomrule
\end{tabular}}
\vspace{-15pt}
\end{table}

\textbf{Robustness to post-processing operations. }
Table~\ref{tab:second detection stage_aug} reports the achieved results on the global test set of the second detection stage, considering post-processing operations that are commonly applied to images. We include JPEG compression with two quality factors, downscaling and upscaling by a factor $2$, and a chain of downscaling and upscaling by a factor $4$.

JPEG compression does not create issues to our detector, which shows close performances to the no editing scenario (see Table~\ref{tab:second detection stage}). The most challenging post-processing is resizing $\times 0.5$, which would need some calibration to maintain high accuracies at fixed thresholding. Indeed, the best \gls{ba} is $91.8\%$, while \gls{ba0} drops by $10$ percentage points and this is reflected in a relatively low \gls{tpr0}. 

This behaviour was not completely unexpected: it has already been shown that downscaling risks to destroy low level artifacts exploited by forensics detectors and to severely affect their performances~\cite{cozzolino2024raising, mandelli2024sar}.
However, it is worth of notice that upscaling reveals helpful to contain this effect, counteracting the loss in performances at the expense of few more false alarms (see the last two columns of Table~\ref{tab:second detection stage_aug}). This ``calibration'' effect of upscaling looks promising and we plan to dedicate it thorough investigations in the near future. 

Apart from these last remarks, it is important to recall that our detector achieves \gls{auc} greater than $0.97$ and best \gls{ba} greater than $91\%$ in all the investigated scenarios. These results prove the robustness of our proposed solution which, despite its simplicity, is highly effective against never seen data and post-processing operations.

\begin{table}[t]
\centering
\caption{Achieved results at the second detection stage for fully synthetic vs laundered (all data), considering post-processing operations applied on images. In italics, the \textit{worst} results per metrics. }
\label{tab:second detection stage_aug}
\resizebox{\columnwidth}{!}{
\begin{tabular}{@{}lccccc@{}}
\toprule
 & JPEG-70 & JPEG-80 & $\times 0.5$ & $\times 2$ & DownUp $\times 4$\\
\midrule
AUC &$0.990$ & $0.991$ & $\mathit{0.970}$ & $0.993$ & $0.976$ \\
$\textrm{B-ACC}_{\max}$ & $95.6\%$  & $95.4\%$ & $\mathit{91.8\%}$ & $96.0\%$ & $92.0\%$ \\
$\textrm{B-ACC}_{@ \textrm{thr}=0}$ & $95.4\%$ & $94.8\%$ & $\mathit{80.2\%}$ & $92.4\%$ & $90.4\%$ \\
$\textrm{TPR}_{@ \textrm{thr}=0}$ & $93.8\%$ & $92.0\%$ & $\mathit{62.2\%}$ & $98.8\%$ & $95.4\%$ \\
$\textrm{FPR}_{@ \textrm{thr}=0}$ & $3.1\%$ & $2.5\%$ & $1.7\%$ & $14.0\%$ & $\mathit{14.5\%}$ \\
\bottomrule
\end{tabular}}
\vspace{-10pt}
\end{table}

%% file: camera_model_problem.tex
\section{Anonymization Effects of Image Laundering: Analysis of Camera Model Identification}
\label{sec:camera_model_problem}

As a final experiment, we investigate the anonymization capabilities of image laundering in the well known forensic problem of camera model identification, i.e.,
identifying the source camera model of a query image.
In the forensic community, it is widely recognized that images taken with the same camera model exhibit a distinct set of artifacts that can differentiate them from images captured by other cameras~\cite{mandelli2022source}. 
Identifying the camera model that produced an image can assist forensic investigators in tracing the original creator of images shared online.

Over the years, the community has developed manifold solutions to attribute an image to its source camera model, considering model-based
and data driven approaches,
even if deep learning-based solutions represent now the standard to deal with such tasks~\cite{mandelli2020training, cozzolino2019noiseprint}.
For instance, a simple end-to-end learning approach was deployed in~\cite{mandelli2020training}, where it is 
shown that standard \gls{cnn}-based detectors can attribute the original camera model with an accuracy greater than $90\%$. 

In this section, we investigate if image laundering can affect performances of state-of-the-art camera model identification detectors. To do so, we test the detector proposed in~\cite{mandelli2020training} over images passed through \gls{sd}-based laundering. 
As pristine data, we use the same test set considered in \cite{mandelli2020training}: images are selected from the well known Vision dataset~\cite{shullani2017vision}; they have a common size of $512 \times 512$ pixels and come from $28$ different camera models. 
With the EfficientNetB0-based detector proposed in the original paper, the achieved camera model identification accuracy on the original test images results $96.15\%$.

Table~\ref{tab:camera_model_accuracy} shows the detection results in case of applying laundering to the test images. Notice the important performance drop: especially for novel \gls{sd} versions, the classification accuracy corresponds to a random guess. 
This counter-forensic effect comes with barely visible traces in the laundered images: if only the laundered copy of the image is available, it would be practically impossible to detect fake generation artifacts by looking at its laundered version. 

This experiment further justifies the need of investigating image laundering as a potential issue for the forensic community. Together with its alarming capability in fooling real-vs-synthetic detectors, image laundering has the potential of undermining the effects of standard forensic methodologies developed to trace back the digital history of an image.

\begin{table}[t]
\centering
\caption{Classification accuracy achieved by the camera model identification detector presented in ~\cite{mandelli2020training}, in absence or presence of SD laundering. In bold, the best accuracy result.}
\label{tab:camera_model_accuracy}
\resizebox{.9\columnwidth}{!}{
\begin{tabular}{@{}lllll@{}}
\toprule
No laundering & SD-1.5 & SD-2.1 & SD-XL & SD-XL-turbo\\
\midrule
$\mathbf{96.15}\%$ & $39.08\%$ & $56.67\%$ & $50.63\%$ & $50.56\%$ \\
\bottomrule
\end{tabular}}
\vspace{-10pt}
\end{table}

%% file: conclusions.tex
\section{Conclusions}
\label{sec:conclusions}

This paper explores the forensic implications of \gls{sd} image laundering, which involves passing an image through \gls{sd} autoencoders to reproduce a synthetic copy of it with high realism and minimal visual artifacts. 
Our experimental campaign shows that image laundering 
has a significant potential to obscure traces of real content, including sensitive and harmful materials that might be mistakenly identified as synthetic, thus compromising the protection of the individuals depicted. 

To combat this issue, we propose a simple yet effective two-stage detection pipeline that reliably distinguishes between pristine, laundered, and fully synthetic images (those generated from text prompts), proving to be a powerful solution under various testing conditions. Eventually, we highlight another concerning aspect of image laundering, which reveals capable to conceal unique artifacts that forensic detectors rely on for camera model identification, thereby severely diminishing their performance. To the best of our knowledge, this is the first paper investigating in depth the laundering issue. We believe our thorough analysis can offer valuable insights to the forensics community, paving the way for more comprehensive understanding and containment of this problem.